\documentclass[lettersize,journal]{IEEEtran}
\usepackage{amsmath,amsfonts}
\usepackage{algorithmic}
\usepackage{algorithm}
\usepackage{array}
\usepackage{textcomp}
\usepackage{stfloats}
\usepackage{url}
\usepackage{verbatim}
\usepackage{graphicx}
\usepackage{cite}
\usepackage{xspace}
\usepackage{multirow}

\ifCLASSOPTIONcompsoc
\usepackage[caption=false, font=normalsize, labelfont=sf, textfont=sf]{subfig}
\else
\usepackage[caption=false, font=footnotesize]{subfig}

\begin{document}
\title{Optimal Energy Management of Plug-in Hybrid Vehicles Through Exploration-to-Exploitation Ratio Control in Ensemble Reinforcement Learning}
\author{Bin Shuai, Min Hua, \emph{IEEE Student Member}, Yanfei Li, Shijin Shuai, Hongming Xu, Quan Zhou*, \emph{IEEE Member}
\thanks{This work was supported in part by Tsinghua University under Grant KF2029 
and in part by Innovate UK under Grant 102253. Corresponding author: Quan Zhou (q.zhou@bham.ac.uk)} 
\thanks{Bin Shuai, Min Hua, Hongming Xu, Quan Zhou are with the Department of Mechanical Engineering, University of Birmingham, Birmingham, B15 2TT, UK.}
\thanks{Yanfei Li, Shijin Shuai are with State Key Laboratory of Automotive Safety and Energy, Tsinghua University, Beijing 100084, China}

}


\maketitle

\begin{abstract}
Developing intelligent energy management systems with high adaptability and superiority is necessary and significant for Hybrid Electric Vehicles (HEVs). This paper proposed an ensemble learning-based scheme based on a learning automata module (LAM) to enhance vehicle energy efficiency. Two parallel base learners following two exploration-to-exploitation ratios (E2E) methods are used to generate an optimal solution, and the final action is jointly determined by the LAM using three ensemble methods. 'Reciprocal function-based decay' (RBD) and 'Step-based decay' (SBD) are proposed respectively to generate E2E ratio trajectories based on conventional Exponential decay (EXD) functions of reinforcement learning. Furthermore, considering the different performances of three decay functions, an optimal combination with the RBD, SBD, and EXD is employed to determine the ultimate action. Experiments are carried out in software-in-loop (SiL) and hardware-in-the-loop (HiL) to validate the potential performance of energy-saving under four predefined cycles. The SiL test demonstrates that the ensemble learning system with an optimal combination can achieve 1.09$\%$ higher vehicle energy efficiency than a single Q-learning strategy with the EXD function. In the HiL test, the ensemble learning system with an optimal combination can save more than 1.04$\%$ in the predefined real-world driving condition than the single Q-learning scheme based on the EXD function.

\end{abstract}


\section{Introduction}
\label{sec:introduction}
\IEEEPARstart{R}{apid} development in transport and mobility systems has a great impact on the environment, such as greenhouse gases (GHS) and other harmful emissions \cite{9847095, liu_review, chen2019comprehensive}. The transportation sector accounts for 20$\%$ of the world's total $CO_{2}$ emission and approximately 71$\%$ \cite{gonul2021electric}, in which off-highway vehicles are responsible for 27$\%$ of $CO_{2}$ emissions in road transport \cite{gregor2018eu}. In this regard, searching for safe, clean, and high-efficiency alternative solutions is the goal for many automobile manufacturers to integrate carbon emission reduction technologies into ICE-based vehicles\cite{tran2020thorough, hua2019hierarchical, chen2019dynamics}. Therefore, electrification of these types of vehicles is in urgent demand, and the hybrid off-highway vehicle is mainstream on the technical roadmap \cite{shuai2022supervisory, hua2020research}. The energy management control strategies (EMSs) for hybrid off-highway vehicles are an indispensable part that coordinates the power flow within the system of multiple power components (e.g., the internal combustion engine) \cite{zhang2019energy}. Much similar work has been done, e.g., for passenger cars, buses, and trucks, and there are two main categories of EMSs, i.e., rule-based methods and optimization-based methods.

Rule-based EMSs highly depend on a set of predefined rules and logic summarized according to the HEV's system characteristics and operation mode, usually developed based on the battery state-of-charge, drive power requirement, and vehicle speed by ``if-else`` logic, to optimize the power allocation between each energy source for achieving the drive requirement and maintaining the battery state-of-charge in an appropriate range \cite{yuan2022optimized}. However, it is not adaptable to different real-world driving cycles due to the predefined rules that need frequent calibrations in the development stages.

Optimization-based methods, including offline and online optimization methods, are necessarily needed to overcome the shortcoming of the rule-based method. For offline optimizations, dynamic programming (DP) is widely used to seek the globally optimal policy based on prior knowledge of the typical driving cycles \cite{lee2020comparative}. Additionally, the EMSs are based on stochastic search, which seeks the optimal control policy by taking the iterative method. These EMSs, including the genetic algorithm (GA) \cite{ding2021design}, simulated annealing algorithm (SA) \cite{hafez2021optimal}, particle swarm optimization (PSO) \cite{mamun2018integrated}, non-dominated sorting genetic algorithm (NSGA-II) \cite{li2021multi} are suitable to address the optimization problems due to their global optimality and robustness. However, the above-mentioned optimal EMSs require an informative prior knowledge of the entire driving conditions, massive computation effort, and less adaptability, and easily fall into the local optimal solutions, which are not able to fully optimize the fuel consumption for the whole trip \cite{sabri2016review}.

The online optimization-based methods usually optimize the hybrid powertrain with limited vehicle information. Model-based predictive control (MPC) is a representative approach that has been widely adopted in the online optimization of HEVs \cite{hou2020hierarchical, chen2023dynamic}. The MPC controller located on the vehicle system runs a rolling optimization process to calculate a series of control signals over a prediction horizon that can achieve the best control performance. And precise predictive models, developed with mathematical functions, are the keys to enabling MPC-based methods to achieve optimal performance in real-world driving \cite{yang2019stochastic}. However, precise models that include detailed real-world dynamics (e.g., aging, changing of drivers) cannot be fully obtained in the development stage \cite{liu2021automated, liu2020vision}. Therefore, learning-based control methods need to be developed to address the limitations of offline and online optimization-based methods.

Learning-based EMSs are promising because it has the potential to break through the limitations of the existing offline and online optimization methods. Among the learning-based methods, reinforcement learning (RL) has gained the most momentum in many research areas, such as driver-behavior classification \cite{lu2018learning, zhao2019identification}, robotics \cite{apolinarska2021robotic}, gaming \cite{liu2021efficient}, and lane detection \cite{xing2020ensemble}. The RL-based EMSs can self-learn to adapt to the different traffic conditions without an accurate model, in which the control parameters of these methods can be updated by continuous interaction with an unknown environment \cite{zou2021dql, yan2022multi}. Liu et al. present several types of research researches for energy management based on the Q-learning method of various topologies of hybrid powertrains, including parallel hybrid powertrain \cite{liu2017reinforcement} and power-split powertrain \cite{liu2019heuristic}. Unlike the MPC-based methods that require an accurate physical model of the vehicle platforms, the RL-based methods are physical model-free. The optimization process is conducted by an RL agent that follows Bellman's optimality theorem to update its knowledge base (a model mapping the state variables and action variables to the expected rewards) through rolling interactions with the environments.

Furthermore, the exploration of new control settings and the exploitation of the existing control policy are two critical procedures in RL \cite{hu2020adaptive}. The RL agent makes a trade-off between exploration and exploitation using the policy with an exploration-to-exploitation ratio (E2E) \cite{carroll2001memory}. If the majority of the experience from exploration is not as good as the best existing returns, over-exploration may impact the learning performance of RL. On the contrary, without enough exploration, the exploitation may lead agents to trap themselves in suboptimal solutions \cite{wang2020reinforcement}. Therefore, a reasonable trajectory of E2E is necessary to achieve a good balance and reduce the bias during the learning process \cite{wang2013backward}. 

Some research is carried out by setting the value of E2E as a constant. For example, Xu et al. implement the Q-learning algorithm with E2E = 0.25 in a mid-size 48V mild parallel HEV \cite{xu2019real}. Qi et al. set the E2E equalling to 0.7 for Q-learning to optimize the fuel consumption in real-time control \cite{qi2015novel}. Liu. et al. present an EMS based on trip information for HEVs using RL with E2E equalling to 0.1 \cite{liu2014power}. In addition, the E2E ratio can be set as a function. The conventional exponential decay (EXD) method has been selected to produce an adaptive E2E trajectory in RL. Zhou et al. proposed multi-step RL and transferred learning-based control methods for hybrid electric vehicles \cite{zhou2019multi}. Shuai et al. propose maximum and random action execution policies to reduce overestimation for standard double Q-learning \cite{shuai2020heuristic}. Other decay methods can also be used to generate the E2E ratio \cite{qi2016data}. However, these methods still suffer from the dilemma of balancing the probability of exploration \cite{xu2020parametric, qi2019deep}, because each decay method has its strengths in the different learning stages and learning rounds.

Ensemble methods show more potential to benefit from all algorithms \cite{krawczyk2017ensemble}, which have been successfully implemented in supervised learning \cite{richman2020nagging} and semi-supervised learning \cite{zhang2013exploiting}. In RL, ensemble methods have been used for representing and learning the value function \cite{sun1999multi}. The critical challenge of the ensemble methods is distributing the proportion between different base learners because all base learners are at the same level of separation power. There hasn't been much research into HEV energy management. One of the first attempts was in \cite{xu2020ensemble}, in which two frameworks of RL-based ensemble methods have been investigated for a power-split HEV. However, this research focuses on the effects of combinations with a different number of vehicle states and different optimization-based methods. The influence of the exploration-to-exploitation (E2E) ratio of the RL-based ensemble method has not been studied yet. Therefore, the RL-based ensemble method with different trajectories of the E2E ratio is worth investigating for in-vehicle applications.

This paper aims to develop a supervisory ensemble system based on a learning automata module (LAM) to optimize energy distribution for a plug-in hybrid vehicle. The main contributions of this paper are shown as follows:

1) Two decay functions, including reciprocal function-based and step-based, are introduced to generate E2E trajectories for the RL agent adaptively.

2) An ensemble learning framework based on a LAM is presented, in which the final decision from all base learners can be executed by the LAM during the learning process through three ensemble methods.

3) Three ensemble learning methods, namely maximum-based, randomly-based, and weighted-based, are illustrated in LAM to determine the final action. Moreover, the different proportions of all base learners using the weighted-based method are explored in this paper.

4) The robustness test for the ensemble learning system is carried out based on the hardware-in-the-loop test platform under four predefined real-world driving cycles.

The remainder of the paper is organized as follows: The system configuration for the studied hybrid electric vehicle is described in Section 2. The supervisor control system with all relevant information required by the RL-based ensemble method is introduced in Section 3, followed by the facilities of the experimental system for evaluation and validation, including the testing driving cycles, software-in-the-loop platform (SiL), and hardware-in-the-loop platform (HiL) in Section 4. Section 5 analyzes the results of supervisory controls with single-agent and ensemble learning systems in both SiL and HiL. Conclusions are drawn in Section 6.

\section{System configuration of connected system}
\begin{figure}[htbp]
\centerline{\includegraphics[width=\columnwidth]{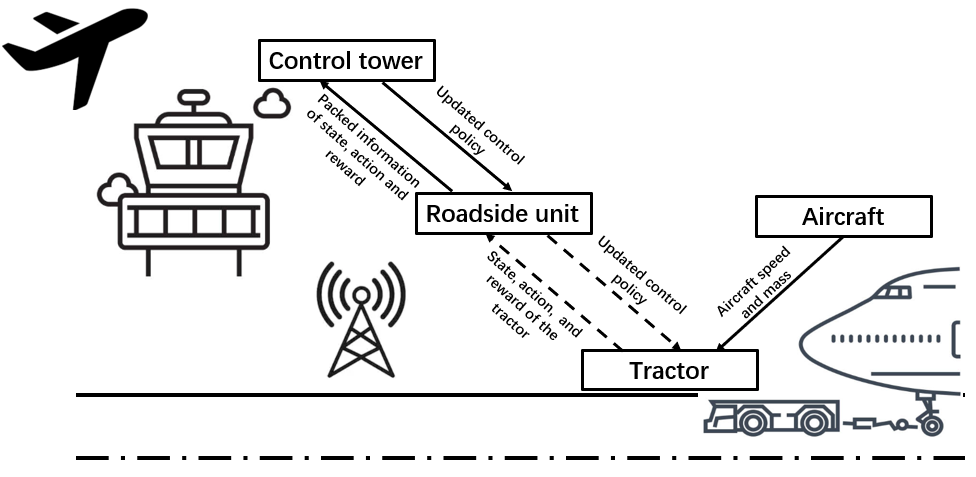}}
\caption{Aircraft-towing operation scenario}
\label{fig0}
\end{figure}

The studied vehicle is a plug-in hybrid electric aircraft-towing tractor, as shown in Fig. \ref{fig0}, which is based on the vehicle specification and operation data provided by the tractor manufacturer. The vehicle works in a particular driving scenario that consists of an aircraft, airport control tower, and V2X (vehicle to everything) communications. The airport control tower sends the command signal to the Roadside Unit (RSU) and receives the vehicle information from RSU through Ethernet. All related vehicle information, including the current vehicle state, engine-generator control signal, and fuel consumption, will be transmitted to the airport control tower (ACT) through the V2X network. The communication between tractor-to-aircraft and the ACT-to-RSU are both enabled by Ethernet. Wi-Fi is used to exchange information between the tractor and RSU.

\begin{figure}[htbp]
\centerline{\includegraphics[width=\columnwidth]{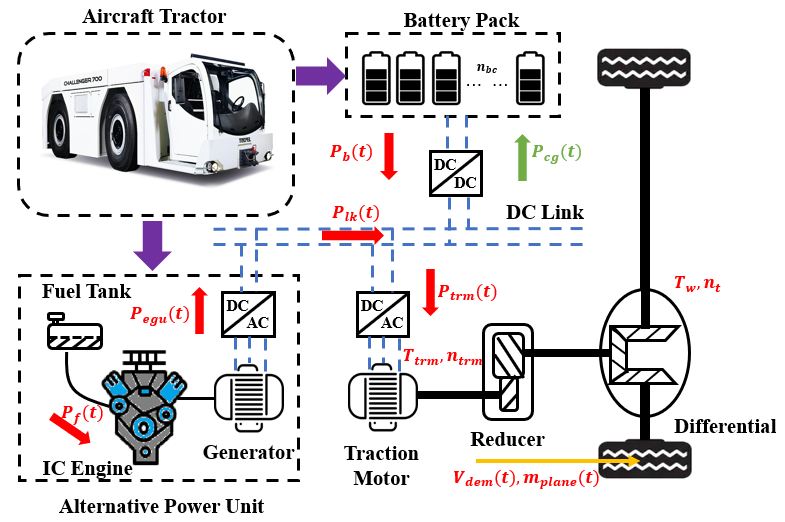}}
\caption{Researched hybrid electric tractor's configuration and power flow}
\label{fig1}
\end{figure}

The aircraft-towing tractor is researched with a series of hybrid powertrain topologies. The internal combustion engine does not have any connection with the wheels, as shown in Fig. \ref{fig1}. The power flow is depicted as follows:

\begin{equation}
\ P_{lk}(t)=P_{apu}(t)+P_{bd}(t)-P_{cg}(t)
\label{eq}
\end{equation}
where $P_{a p u}(t)$ is alternative power unit (APU) output power, the battery pack (BP) discharge power is $P_{b d}(t)$; the charge power of BP is $P_{cg}(t)$ ; 
and the DC-link power is which can be described as follow:
\begin{equation}
\ P_{lk}(t)=P_{trm}(t)+P_{\text {loss, trm }}(t)
\label{eq}
\end{equation}
where $P_{\text {loss, trm }}(t)$ is the traction motor power loss, and the power for running the aircraft-towing tractor is $P_{trm}(t)$. The key vehicle specifications are summarized in Table \ref{table1}. The details in the modeling of the components are described as follows:

\subsection{Traction motor}

In this research, the traction motor is provided by TM4 electrodynamic system Ltd. It has a nominal power of 245kW. As for the studied aircraft-towing tractor, the tractor generally works in low-speed and high-load driving conditions, which means the regenerative braking system is not cost-efficient. Therefore, the traction motor is only operating in traction mode. It can be modeled as:

\begin{equation}
\ P_{trm}(t)=\frac{T_{trm} \times n_{trm}(t)}{9550}
\label{eq}
\end{equation}
where the $P_{t r m}$ is the output power from the traction motor; $T_{trm}$ and $n_{trm}$ are the torque and rational speed of the traction motor. And the tractor motor loss can be calculated by a quadratic function of the motor torque. The efficiency of the traction motor is described as follows:

\begin{equation}
\ \eta_{t r m}(t)=\frac{P_{t r m}\left(T_{\text {trm }}(t), n_{t r m}(t)\right)}{P_{\text {trm }}\left(T_{\text {trm }}(t), n_{t r m}(t)\right)+P_{\text {loss,trm }}\left(T_{t m}(t), n_{t r m}(t)\right)}
\label{eq}
\end{equation}
where $P_{loss}$ is the power loss of the traction motor, which can also be described as a function of motor torque. The function can be formed as a quadratic function in this paper using the data provided by the motor supplier.

\begin{table}[]
\centering
\caption{Aircraft-towing tractor profile}
\label{table1}
\begin{tabular}{cc}
\hline
Specifications        & Value     \\ \hline
Aircraft-tractor mass & 16 tonnes \\
Front area            & 6.8 m2     \\
Drag coefficient      & 0.8       \\
Friction coefficient  & 0.02      \\
Fixed gear ratio      & 25        \\ 
Fixed gear ratio      & 25        \\ \hline
\end{tabular}
\end{table}

\subsection{Engine-generator unit}

The engine generator (EGU) of the aircraft-towing tractor is provided by JCB company. It compromises an 86.20kW diesel engine and a three-phase motor generator. The power provided by EGU can be modeled as a quadratic function \cite{hu2013energy}.
\begin{equation}
\ P_{ef}=b_{2} \times P_{egu}^{2}+b_{1} \times P_{egu}+b_{0}
\label{eq}
\end{equation}
where he coefficient $b_{1}$ and $b_{2}$ are decided with curve fitting using the fuel rate data from Table \ref{table2}. The $P_{e f}$ is the equivalent power fuel consumed, which can be evaluated by:

\begin{equation}
\ P_{ef}=\frac{\dot{v}_{f} \rho_{f} H_{f}}{3600}
\label{eq}
\end{equation}
where ${v}_{f}$ rate of fuel consumption in L/h; $\rho_{f}=0.87kg/L$ is the fuel density; ${H}_{f}=44\times 10^{6}J/kg$ is the heat value of the fuel. The energy efficiency of the EGU can be calculated as:

\begin{equation}
\ \eta_{f2e}(t)=\frac{P_{\text {egu}}(t)}{P_{\text {fuel}}\left(P_{\text {egu}}(t)\right)}
\label{eq}
\end{equation}
where $P_{egu}$ is EGU output power. The main parameters of the function $P_{fuel}$ are calibrated by calculating the real-time $P_{fuel}(t)$ with the measured real-time fuel consumption rate $\dot{m}_{f}$.

\begin{table}[]
\centering
\caption{Engine-generator unit profile}
\label{table2}
\begin{tabular}{cc}
\hline
Specifications       & Value      \\ \hline
Maximum power        & 86.20 kW   \\
Frequency            & 50 Hz      \\
Voltage              & 230V$\sim$ \\
Phase                & 3-AC       \\
Fuel type            & Diesel     \\
50\% load fuel rate  & 13.00 L/h  \\
75\% load fuel rate  & 18.60 L/h  \\
100\% load fuel rate & 24.10 L/h   \\
Fuel tank capacity   & 285 L      \\ \hline
\end{tabular}
\end{table}

\subsection{Battery package}
The battery pack (BP) contains 8200 battery cells, type NCR-18650 series, provided by Panasonic Automotive $\&$ Industrial System Ltd. The changing rate of the battery's state-of-charge (SoC) can be evaluated as follows:
\begin{equation}
\ {S\dot{o}C} =I_{batt}(t) / C_{bat}
\label{eq}
\end{equation}
where $C_{bat}$ is the nominated capacity of the battery cell, and its value is 2.450Ah, and $I_{batt}$ is the cell current, which can be determined by:
\begin{equation}
\ I_{\text {batt}}(t)=\frac{P_{batt}(t)}{U_{batt}(t) \cdot N u m_{\text {batt}}}
\label{eq}
\end{equation}
where $P_{batt}$ is the power of battery pack; $Num_{batt}$ is the number of battery cell and $U_{batt}$ is the voltage of the battery pack.

\section{Supervisory control system based on ensemble learning methods}
\subsection{Synchronized-based ensemble learning system}
\begin{figure*}[htbp]
\centerline{\includegraphics[width=0.8\textwidth,height=0.5\textwidth]{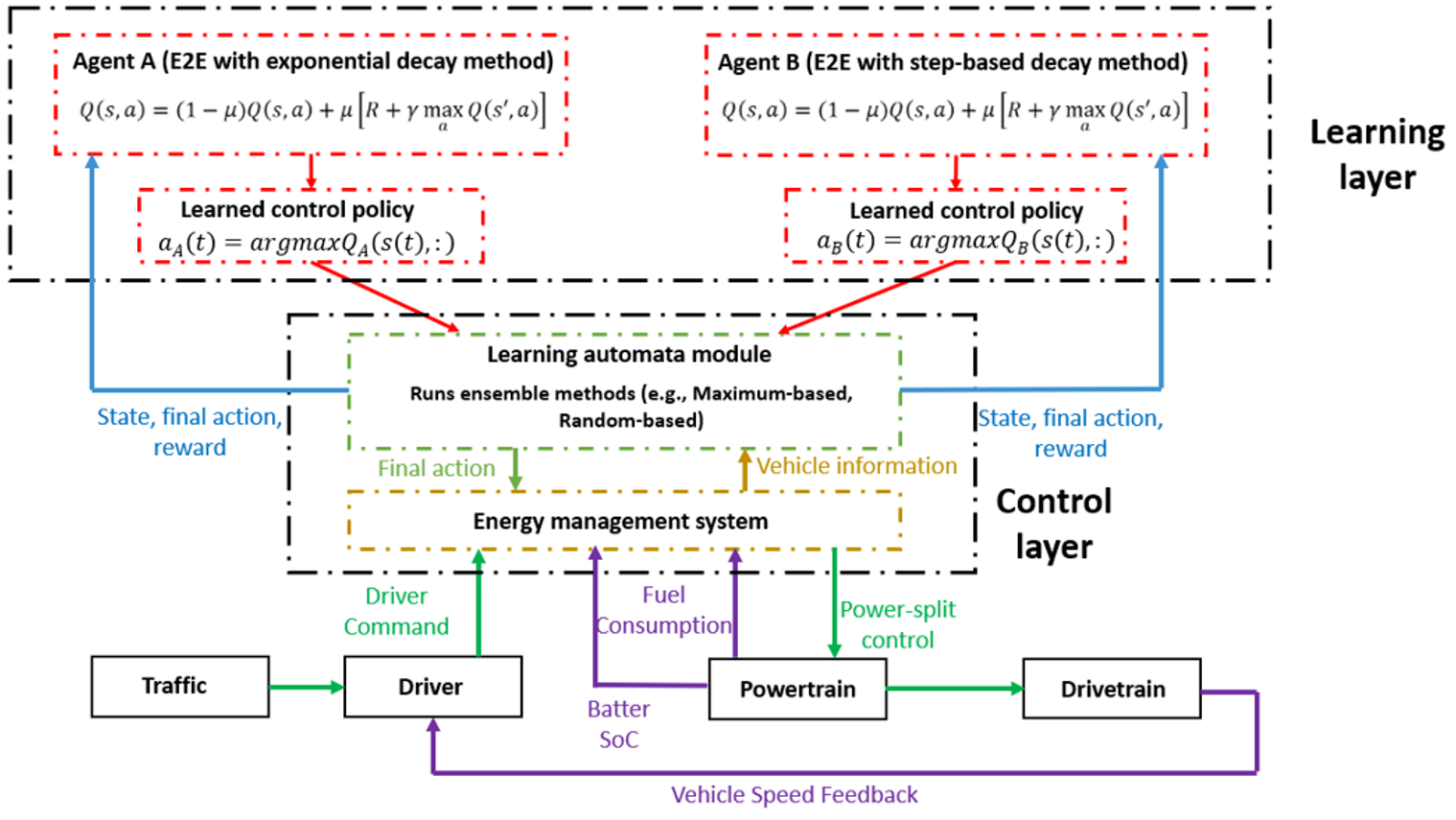}}
\caption{The framework of synchronized-based ensemble learning system}
\label{fig2}
\end{figure*}

The synchronized-based ensemble learning system's procedure is first described in Fig. \ref{fig2}, consisting of the learning and control layers. The vehicle information (e.g., driver's power demand, battery state-of-charge, control policies) will be transmitted between two layers through the V2X (vehicle to everything) network. Especially, the vehicle's velocity and acceleration states can be accurately determined by integrating its kinematics and dynamics using the consensus Kalman filter, as demonstrated in \cite{xia2022autonomous}.
Moreover, in the learning layer, two parallel reinforcement learning agents with different exploration-to-exploitation (E2E) ratio policies are developed to explore the unknown driving conditions and update learned experience in its knowledge bases (Q-tables). Then in the control layer, a learning automata module is used to determine the final engine-generator signal by taking the ensemble methods. 

This framework can take advantage of the ensemble during the learning stage, in which a learning process and ensemble process are running synchronously. Moreover, the LAM determines a superior final action to the target vehicle for better performance using three ensemble methods. The detailed working step is described as follows: 

\textbf{Step 1}: The two parallel agents are located in the learning layer, in which agent A runs the exponential decay method and agent B runs the step-based decay method. At the time step , the energy management system collects vehicle information and transmits it to the LAM. Then the two base learners take the observation with current vehicle states, $s(t)$, and vote its optimal action$a_{A}(t)$ and $a_{B}(t)$. Agent A follows the exponential decay method, the $a_{A}(t)$ is defined as below: 

\begin{equation}
\ a_{A}(t)=\left\{\begin{array}{lr}
\operatorname{argarg} \max _{a(t) \in A}\left[Q^{A}(s(t),:)\right] & \text { if } T \geq \theta \\
\text { randomly pick } & \text { if } T<\theta
\end{array}\right.
\label{eq}
\end{equation}

And the agent B follows the step-based decay method, the action $a_{B}(t)$ is presented as follows:

\begin{equation}
\ a_{B}(t)=\left\{\begin{array}{lr}
\operatorname{argarg} \max _{a(t) \in A}\left[Q^{B}(s(t),:)\right] & \text { if } T \geq \theta \\
\text { randomly pick } & \text { if } T<\theta
\end{array}\right.
\label{eq}
\end{equation}

\textbf{Step 2}: In LAM, the final action is determined by taking the ensemble methods, including maximum value-based, random-based, and weighted-based. The description for each ensemble method is described as follows:

\subsubsection {Maximum-based method}

For the maximum-based method, the final action is picked up with the highest merit-function value between the actions from two base learners by:

\begin{equation}
\ \pi_{LA}: a_{final}(t)=\arg \max _{a(t) \in U}\left[a_{A}(t), a_{B}(t)\right]
\label{eq}
\end{equation}
where $a_{A}(t)$and $a_{B}(t)$ are two actions picked from each base learner with maximum value in current vehicle state $s(t)$; $\pi_{LA}$is the execution policy of learning automata for determining the final action.

\subsubsection{Random based method}
A deciding variable is introduced to compare with a random comparing variable. The final action is determined using the comparison result:

\begin{equation}
\ \pi_{L A}: a_{\text {final }}(t) \leftarrow\left\{\begin{array}{ll}
a_{A}(t) & \text { if } Y \geq T \\
a_{B}(t) & \text { if } Y<T
\end{array}\right.
\label{eq}
\end{equation}
where $Y$ is the random comparing variable between 0 and 1; $T$ is the deciding variable that determines the final action between two base learners. If $ Y\geq T$, the final action $a_{final }(t)$ is selected from the base learner A. Otherwise, an action $a_{final }(t)$ is collected from base learner B.

\subsubsection{Weighted-based method}
The weighted-based method considers the advantages of different E2E ratio policies from agent A and agent B. The final action is determined by taking the consideration each agent's proportion, which can be calculated by

\begin{equation}
\ a_{\text {final }}=\mu a_{A(t)}+\delta a_{B}(t)
\label{eq}
\end{equation}
where $\mu$ and $\delta$ are the coefficient factors determining the proportions in action $a_{A(t)}$ and action $a_{B(t)}$, respectively.

\textbf{Step 3}: The final action is conducted to the engine generator through the energy management system, and the powertrain feeds back the LAM with a reward R. All the related information is uploaded to the learning layer. The control policies for each base learner are used in this final action to update its knowledge base by:

\begin{equation}
\begin{array}{l}
\begin{split}
    Q_{A}\left(s(t), a_{A}(t)\right) \leftarrow Q_{A}\left(s(t), a_{A}(t)\right) \\ 
    + \left[R+maxQ_{A}(s(t+1),:)-Q_{A}\left(s(t), a_{\text {final }}\right)\right] \\
    Q_{B}\left(s(t), a_{B}(t)\right) \leftarrow Q_{B}\left(s(t), a_{B}(t)\right) \\
    +\left[R+maxQ_{B}(s(t+1),:)-Q_{B}\left(s(t), a_{\text {final }}\right)\right] 
\end{split}  
\end{array}
\label{eq}
\end{equation}
where $maxQ_{A}(s(t+1),:)$ and $maxQ_{B}(s(t+1),:)$ are the maximum value from base learner A and base learner B at vehicle state $s(t+1)$.

\subsection{Decay functions for exploration-to-exploitation ratio}
In conventional Q learning algorithms, exploration and exploitation are controlled based on the comparison between a random number, $\varepsilon $, with the value of an explorational decay function, $\theta =\alpha _{1}^{K} $, where $K$ is the current learning iteration, $\alpha _{1} $ is the initial value of E2E; If $\varepsilon > \theta $, the algorithm will exploit based on the current control policy. Otherwise, it will explore possible better control policies by randomly picking up an action value. To study the Q-learning performance with different E2E, this paper introduces two new decay functions, including step-based decay and reciprocal function based, to generate optimal action.

\subsubsection{Step-based decay policy}

One popular learning rate scheduler is step-based decay, where we systematically drop the learning rate after specific epochs during training. The step decay function can be seen as a piecewise function. In other words, the learning rate or E2E is constant for a number of epochs, then drops, is constant once more, then drops again, etc. The basic step-based decay function can be defined as below:

\begin{equation}
\ \theta = \alpha _{1} \times F^{round\left ( \frac{1 + E}{D}  \right ) } 
\label{eq}
\end{equation}
where $\alpha _{1}$is the initial learning rate (Usually initial learning rate = 0.5 and the E2E = 0.8). $F$ is a factor value controlling the rate at which the learning rate drops; the larger factor $F$ is, the slower the learning rate or E2E will decay. Conversely, the smaller the factor $F$, the faster the learning rate or E2E will decay. $D$ is the scale factor (typically the $\theta$ decreases after ten times of learning) that controls how many learning iterations the $\theta$ drops;  is the current learning iteration.

\subsubsection{Reciprocal function-based decay policy}

The mechanism of the reciprocal function-based decay method is very similar to the stochastic gradient descent method (SGD), which is widely used for efficient deep learning network training, as illustrated in \cite{liu2022yolov5}. It drops the learning rates and E2E faster than the exponential decay and step-based decay method. The reciprocal decay is described as follows:

\begin{equation}
\ \theta = \frac{\alpha _{1} }{1+  R_{decay}\ast D } 
\label{eq}
\end{equation}
where $R_{decay}$ is a decay rate, if the $R_{decay}=0$, there is not influence on learning rate or E2E. If the $R_{decay}$ is specified, which will decrease $\theta$ by the given fixed amount.

\subsubsection{Learning automata module}
The learning automata module (LAM) is a classifier that identifies the best final action by responding from the unknown environment after picking different actions at different times. The interaction of learning automata with the environment The LAM selects an action from all base learners and evaluates this action with a merit function. The knowledge bases of all base learners are updated by repeating this process, and finally, the optimal action is gradually identified. The global solution can be fined among all base learners. There are three essential components in LAM that can be formally modeled as:

\begin{equation}
\ LA=\left \{ S,A,R \right \} 
\label{eq}
\end{equation}

The $S$ is the current vehicle state which is responsible for determining the current state of the vehicle system based on the sensor signal, and the driver power demand and battery state-of-charge are selected \cite{shuai2020heuristic}, which is defined as follows:

\begin{equation}
\ s\left ( t \right ) =\left [ P_{dem}(t),SoC(t)  \right ] 
\label{eq}
\end{equation}
where $s(t)\in S$ is the current state at the $t^{th} $ time step; $P_{dem} (t)\in\left \{  0kW\le P_{dem}\le 253kW\right \}$ is the driver's power demand value at the $t^{th} $ time step; $SoC(t)\in \left \{ 20\%\le SoC\le 80\% \right \} $ is the battery SoC value at the $t^{th} $ time step.

The $A$ is a set of actions of LAM, which will be used to control the power rate of the engine-generator; The execution policy $\pi _{exe} $ is used to determine the action $a(t)\in A$ based on the current vehicle state $a(t)\in A$, which is defined below: 

\begin{equation}
\ a(t)\gets \pi _{exe} \left ( Q (s(t),A)\right ) 
\label{eq}
\end{equation}

The $R$ is used to evaluate the selected action, which is consisted of a merit function. The merit function is defined as:

\begin{equation}
r(t)=\left\{\begin{array}{c}
\mathrm{r}_{ini}-P_{\text {loss }}(l) \mathrm{SoC}(L) \geq S_{\mathrm{SoC}}{ }_{ref} \\
\mathrm{r}_{ini}-P_{\text {loss }}(t)-\mu\left|S_0-C_p-\mathrm{SoC}(t)\right| \operatorname{SoC}(t)<\mathrm{SoC}_{ref}
\end{array}\right.
\end{equation}
where $SoC_{ref} $ is the reference battery SoC value that is chosen to maintain the battery SoC within an acceptable range (for the best performance and health of the battery $SOC_{ref} $ = 28$\%$). $\alpha $ is a scale factor to balance the consideration of battery SoC level and power efficiency; $P_ {loss }(t)=Loss_{eng} (t)+Loss_{batt} (t)$ is the total power loss of t engine $Loss_{eng} (t)$ and battery; the power loss of engine and power loss of battery $Loss_{batt} (t)$ can be calculated by:

\begin{equation}
\ \left.\begin{array}{l}
L_{\text {eng }}(t)=\dot {m}_{f}(t)\cdot H_{f}-\frac{T_{\text {eng }}(t) \cdot n_{\text {eng }}(t)}{9550} \\
L_{\text {batt }}(t)=R_{\text {loss }}(S o C) \cdot I_{\text {batt }}(t)^{2}
\end{array}\right\}
\label{eq}
\end{equation}
where $\dot m_{f} $ is the fuel rate in real-time. $T_{eng} $ and $n_{eng} $ are the torque and running speed of the diesel engine, respectively; $l_{batt} $ is BP's current; the equivalent internal battery resistant $R_{loss}$ is a function of battery SoC; and diesel fuel heat value is $H_{f}=44\times 10^{6} J/kg$.

\section{Experimental results and discussion}
\subsection{Experimental design}
The two proposed E2E methods are firstly run in the software-in-the-loop test platform (SiL) with the initial value E2E = 0.8 with an initial battery state-of-charge of 50$\%$ and 125 rounds of learning. The learning performance is monitored through vehicle energy efficiency under predefined driving cycle one. The results of the two proposed E2E methods are compared to the conventional exponential-based decay methods. Next, the evaluation of the ensemble learning system is carried out with three ensemble methods. Two rounds of the hardware-in-the-loop test are validated. One round is for the real-time feasibility of the ensemble learning system to validate in the hardware-in-the-loop testing platform. Another one is for the robustness test of the ensemble learning system under the other three predefined real-world driving cycles. Moreover, the results of the HiL test are compared to the conventional exponential decay method.

\subsubsection{Driving cycles}

Four predefined driving cycles provided by the tractor manufacturer are selected to evaluate the real-time performance of proposed policies. The power requirement of each driving cycle is shown in Fig.4. One driving cycle is set as a learning cycle for machine learning, and the other three are used to evaluate the real-time performance of model-free supervisory controls.

\begin{figure}[htbp]
\centerline{\includegraphics[width=\columnwidth]{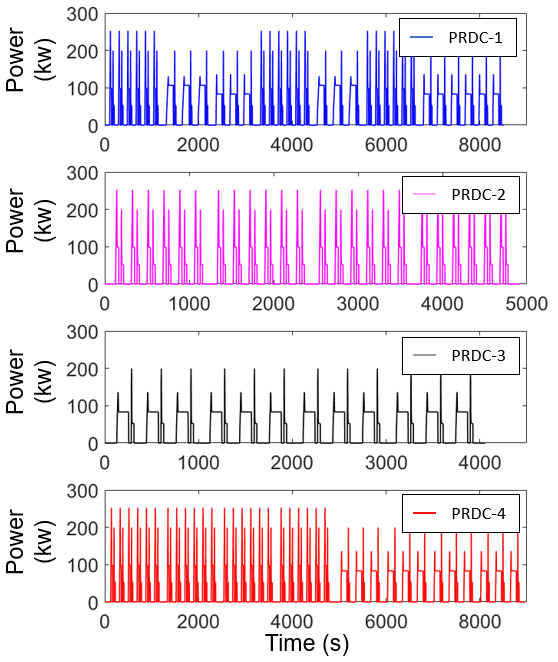}}
\caption{Power of the predefined driving cycle}
\label{fig3}
\end{figure}

\subsubsection{Hardware-in-the-loop testing platform}

The feasibility of model-free supervisory control methods will be validated through the hardware-in-the-loop testing platform (HiL), developed based on the ETAS's vehicle experiment facilities, as shown in Fig.5. The working step for HiL is: 1) the control strategy is programmed in the development computer with MATLAB/Simulink. 2) The control strategy will then be loaded into the ETAS IP and compiled into C-code through ETAS' Experimental Environment software. 3) The complied control strategy will be flashed into the DESK-LABCAR via Ethernet. 4) The vehicle performance can be observed and recorded by ETAS Experimental Environment.

\begin{figure}[htbp]
\centerline{\includegraphics[width=\columnwidth]{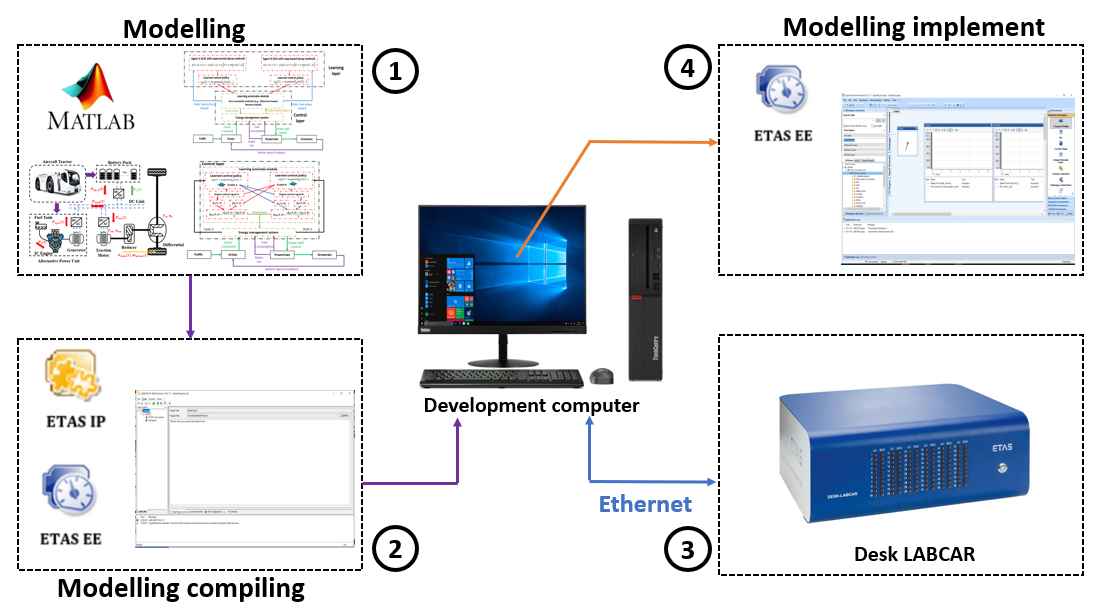}}
\caption{The hardware-in-the-loop platform for real-time testing.}
\label{fig4}
\end{figure}

\subsection{Learning performance of single-agent with the step-based decay and reciprocal function-based decay methods}

The learning performance of the step-based decay (SBD) method, reciprocal function-based decay (RBD) method, and exponential decay (EXD) method in SiL was demonstrated in Fig.6. In order further to analyze the difference between the three decay methods, the E2E trajectories of the three methods are shown in Fig.7. The SBD method performed the best among the three methods, with 1.69$\%$ and 0.72$\%$ higher than the EXD and RBD methods, at the end of the learning. Compared to the three E2E trajectories, the E2E of the SBD dropped mildly, which still had tremendous potential to explore new engine control signals between the 20th and 60th learning iterations. The E2E of the EXD method and the RBD method decreased rapidly. These two methods did not leave enough time for the agent to understand the unknown driving conditions. Additionally, the SBD method achieved the highest improvement rate of 6.21$\%$ (final vehicle energy efficiency compared with initial vehicle energy efficiency), higher than the RBD method with 1.36$\%$ and 0.33$\%$ than the EXD method, respectively.

\begin{figure}[htbp]
\centerline{\includegraphics[width=\columnwidth]{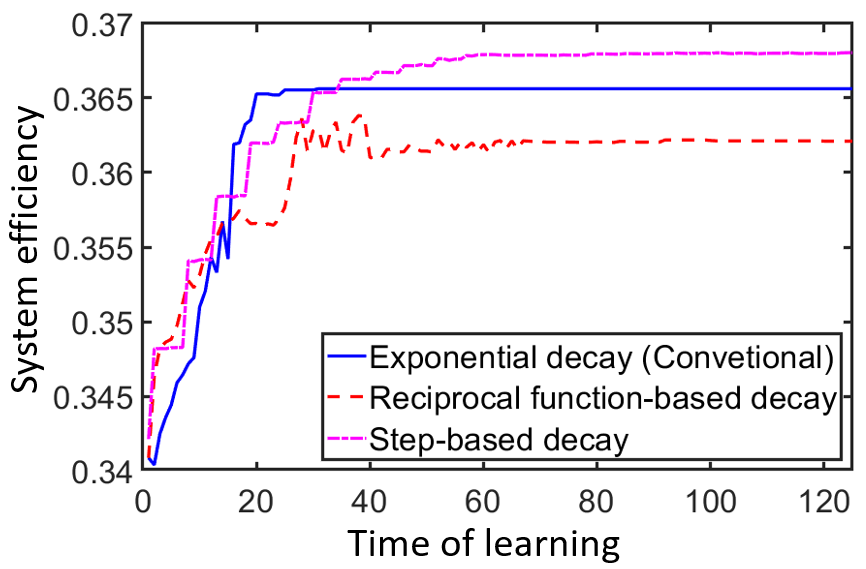}}
\caption{The variation curve of the deciding factor}
\label{fig5}
\end{figure}

\begin{figure}[htbp]
\centerline{\includegraphics[width=\columnwidth]{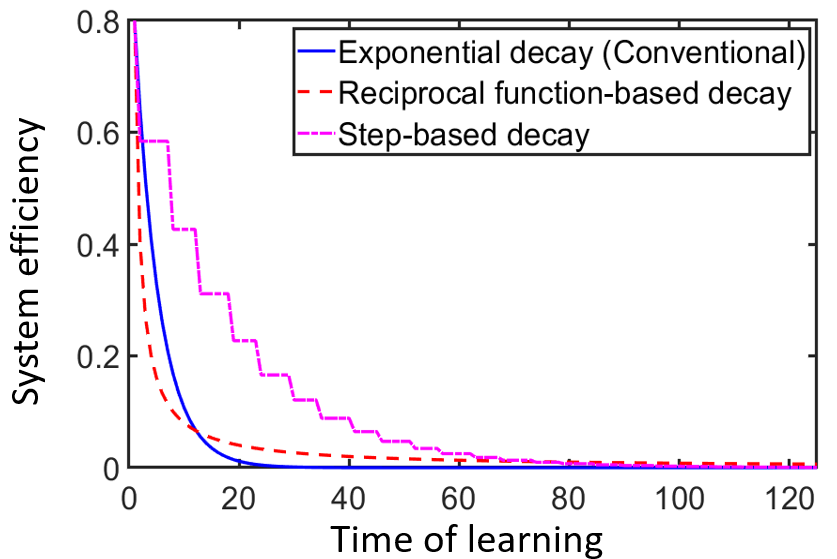}}
\caption{The variation curve of the deciding factor}
\label{fig6}
\end{figure}

\subsection{Improvement in energy efficiency with the ensemble learning system}
Firstly, the investigation of vehicle energy efficiency was obtained by three ensemble methods, including the random-based method, maximum-based method, and weighted-based method, to select the most effective ensemble method that achieves the highest vehicle energy efficiency at the end of learning. Secondly, the weighted-based ensemble method was studied with internal characteristics to explore the best performance, and the results were compared to the single-agent methods (SBD, RBD, and EXD). The experiment mentioned above was conducted by 125 times of learning under predefined driving cycle one with an initial battery SoC of 50$\%$ based on the software-in-the-loop test platform.

\subsubsection{Learning performance of ensemble learning system with different competition methods}
The learning performance of three ensemble methods, including random-based, maximum-based, and weighted-based methods, are shown in Fig.8. The energy efficiency obtained by the maximum-based and random-based methods is unstable and convergent very slowly throughout the learning process. Moreover, the vehicle energy efficiency obtained by the weighted-based method improved smoothly and achieved the highest energy efficiency at the end of the learning process, up to 1.06$\%$ and 5.06$\%$ higher than the maximum-based method and random-based method, respectively. The weighted-based method can provide more diversity of the engine control signal by taking different proportions from all base learners. Usually, the default proportion of each learner is 50$\%$, which assumes that the final action was determined by half of the base learner A and half of base learner B. Therefore, the impact of different proportions of two learners is necessary to study in the weighted-based method. The investigation was conducted by changing the proportion from 10$\%$ to 90$\%$ of each learner. The test was conducted by 125 rounds of learning under predefined driving cycle one with an initial battery state-of-charge of 50$\%$. Each test with a different proportion of learner A and learner B was repeated 25 times.

\begin{figure}[htbp]
\centerline{\includegraphics[width=\columnwidth]{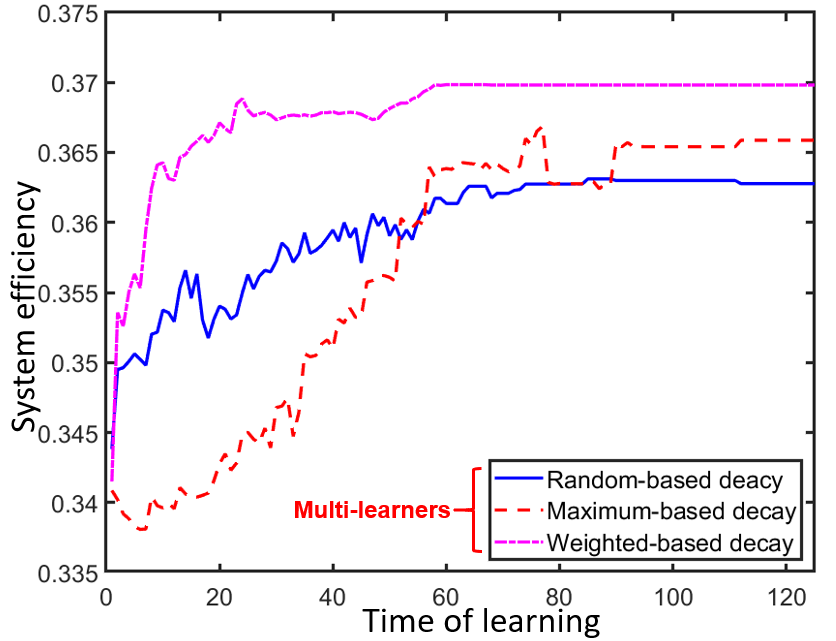}}
\caption{Learning performance of three ensemble methods}
\label{fig7}
\end{figure}

The results of different proportions for two learners using the weighted-based method were shown in Table.3, in which learner A followed the SBD method, and learner B followed the EXD method. The vehicle's energy efficiency achieved the highest of 36.98$\%$ when learner A represented 10$\%$, and learner B represented 90$\%$.

\begin{table}[]
\centering
\caption{Different proportions of two base learners}
\label{table}
\begin{tabular}{ccc}
\hline
Agent A (SBD) & Agent B (EXD) & Vehicle's energy efficiency \\ \hline
10\%          & 90\%          & 36.80\%                     \\
20\%          & 80\%          & 36.59\%                     \\
30\%          & 70\%          & 36.80\%                     \\
40\%          & 60\%          & 36.84\%                     \\
50\%          & 50\%          & 36.51\%                     \\
60\%          & 40\%          & 36.68\%                     \\
70\%          & 30\%          & 36.85\%                     \\
80\%          & 20\%          & 36.85\%                     \\
\textbf{90\%} & \textbf{10\%} & \textbf{36.98\%}            \\ \hline
\end{tabular}
\end{table}

\subsubsection{Level of learning with the single-learner and multi-learners}
The learning performance of the ensemble learning system using a weighted-based method (90$\%$ of learner A and 10$\%$ of learner B) is compared to results obtained by the single-learner-based supervisory control system (SBD, RBD, EXD). The results are shown in Fig.9. The weighted-based method achieved 1.15$\%$ higher vehicle energy efficiency than the baseline method (exponential-decay method). Moreover, at the end of learning, the weighted-based method outperformed all single-learner methods by at least 0.5$\%$ vehicle energy efficiency. The whole learning process can be divided into two stages. In the early learning stage (before the 60th time of learning), the weighted-based method took the leading, consistently higher than single-agent methods. Because the learning automata take the strength 1) of the SBD method in high probability to explore the new attempts; 2) and taking 'suggestion' of EXD to generate more new attempts of an engine control signal. Then, in the latter learning stage (after the 60th time of learning), all the methods slow down the improvement after a 'knee point. This is because the E2E is very close to zero, which leads the LAM to determine the final action from the existing knowledge base.

\begin{figure}[htbp]
\centerline{\includegraphics[width=\columnwidth]{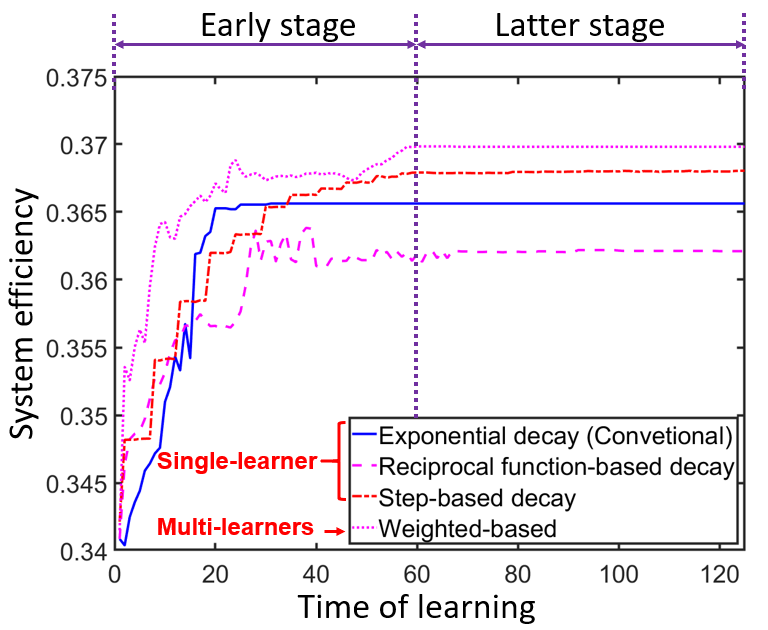}}
\caption{The learning performance of four model-free control methods}
\label{fig8}
\end{figure}

\subsection{Real-time performance of ensemble reinforcement learning system}

The ensemble learning system with the weighted-based method outperforms the other two methods. Therefore, it needs to validate the real-time feasibility based on the hardware-in-the-loop test platform. The real-time performance of the weighted-based method is carried out under predefined driving cycle one with an initial battery SoC of 50$\%$ and compared to the exponential-based method results. The results are demonstrated in Fig.10. We compare the real-time data, including the total energy loss of the vehicle, the variation of the battery state-of-charge, and the equivalent energy loss in the battery and engine generator, in each subfigure. The proposed method outperformed the conventional exponential decay method by making more reductions in total energy loss. This can be seen from the trajectory of battery SoC. The proposed method can maintain the battery SoC closely to the predefined threshold (). This led to significant energy savings in the battery with the proposed method.

\subsection{Robustness testing of ensemble reinforcement learning methods}

In actual driving situations, the hybrid electric vehicle energy management system will encounter various driving conditions. Therefore, a robustness test is required for the proposed ensemble learning system. Three predefined driving cycles (PRDC-2 to 4) are selected to simulate unknown real-world driving conditions. The 'End SoC' (battery state-of-charge after running a whole driving cycle) and 'OEC' (overall energy usage after running a whole driving cycle) are given in Table 4. The ensemble learning with the weighted-based method (WBD) is compared to the baseline method with the exponential decay method (EXD). The rate of total energy savings can be calculated as follow:

\begin{equation}
\ \Delta =\frac{Ef_{EXD}-Ef_{WBD}  }{Ef_{EXD}} 
\label{eq}
\end{equation}
where $\Delta$ is the rate of energy saving for each model-free method; $Ef_{EXD}$ is the vehicle operation total energy cost using the EXD method; $Ef_{WBD}$ is the total energy cost of the vehicle operation using the WBD. In order to keep the battery pack for long-life cycle safety. We put a threshold; when the battery State-of-charge is below 28$\%$, the engine generator will automatically get involved to charge the battery pack. Compared to the EXD, the WBD can save at least 1.04$\%$ energy. The highest overall energy-saving rate of WBD is 2.61$\%$.

\begin{figure}[htbp]
\centerline{\includegraphics[width=\columnwidth]{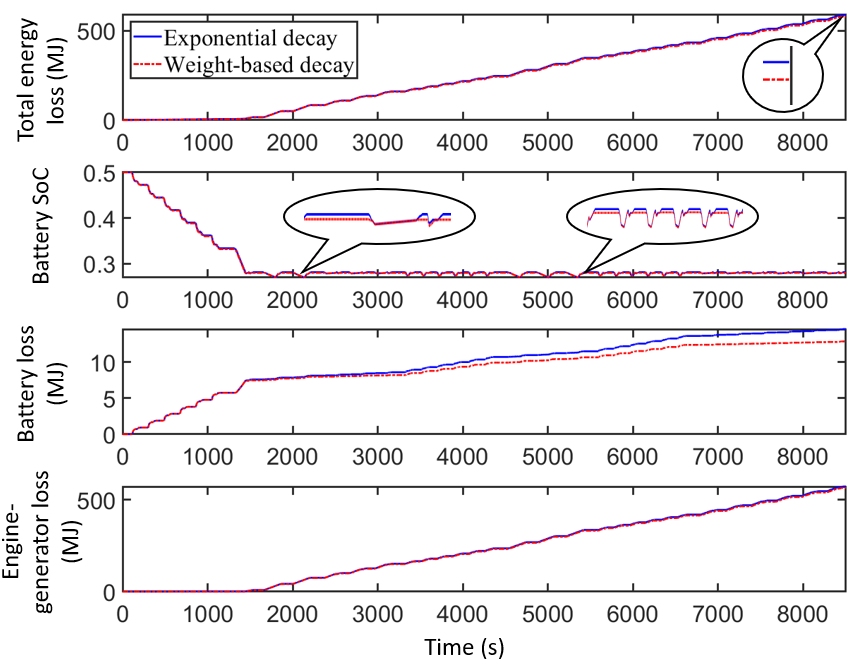}}
\caption{Real-time performance of three model-free methods}
\label{fig9}
\end{figure}

\begin{table}[]
\centering
\caption{Different proportions of two base learners}
\label{table}
\begin{tabular}{ccc}
\hline
Agent A (SBD) & Agent B (EXD) & Vehicle's energy efficiency \\ \hline
10\%          & 90\%          & 36.80\%                     \\
20\%          & 80\%          & 36.59\%                     \\
30\%          & 70\%          & 36.80\%                     \\
40\%          & 60\%          & 36.84\%                     \\
50\%          & 50\%          & 36.51\%                     \\
60\%          & 40\%          & 36.68\%                     \\
70\%          & 30\%          & 36.85\%                     \\
80\%          & 20\%          & 36.85\%                     \\
\textbf{90\%} & \textbf{10\%} & \textbf{36.98\%}            \\ \hline
\end{tabular}
\end{table}

\begin{table}[]
\centering
\caption{Performance of the CBD, FDB and LBD under predefined driving conditions }
\label{table}
\begin{tabular}{cccccc}
\hline
Driving Cycle          & Initial SoC & Method & End SoC & OEC (MJ) & Savings \\ \hline
\multirow{6}{*}{PRDC-2}
                        & 30\%        & EXD    & 28.13\% & 333.98   & -       \\
                        & 30\%        & WBD    & 28.03\% & 327.51   & 1.94\%  \\
                        & 50\%        & EXD    & 28.13\% & 239.25   & -       \\
                        & 50\%        & WBD    & 28.02\% & 234.38   & 2.03\%  \\
                        & 70\%        & EXD    & 28.13\% & 142.07   & -       \\
                        & 70\%        & WBD    & 28.01\% & 138.36   & 2.61\%  \\ \hline
\multirow{6}{*}{PRDC-3}                  
                        & 30\%        & EXD    & 28.13\% & 295.67   & -       \\
                        & 30\%        & WBD    & 28.02\% & 292.03   & 1.23\%  \\
                        & 50\%        & EXD    & 28.13\% & 201.75   & -       \\
                        & 50\%        & WBD    & 28.01\% & 199.12   & 1.31\%  \\
                        & 70\%        & EXD    & 28.13\% & 105.11   & -       \\
                        & 70\%        & WBD    & 28.03\% & 103.37   & 1.65\%  \\ \hline
\multirow{6}{*}{PRDC-4}
                        & 30\%        & EXD    & 28.13\% & 395.41   & -       \\
                        & 30\%        & WBD    & 28.02\% & 391.31   & 1.04\%  \\
                        & 50\%        & EXD    & 28.13\% & 305.05   & -       \\
                        & 50\%        & WBD    & 28.02\% & 301.11   & 1.31\%  \\
                        & 70\%        & EXD    & 28.13\% & 211.09   & -       \\
                        & 70\%        & WBD    & 28.03\% & 207.43   & 1.76\%  \\ \hline
\end{tabular}
\end{table}

\section{Conclusion}

This paper studied a new energy-efficiency supervisory ensemble learning system for an off-highway vehicle. An ensemble learning framework is proposed based on a learning automata module (LAM). The LAM follows three ensemble learning methods, including maximum-based, random-based, and weighted-based methods, to make the final control signal to the engine. Two decay functions, 'Reciprocal function-based decay' (RBD) and 'Step-based decay' (SBD), are proposed based on the exponential decay (EXD) function to generate E2E trajectories for two base learners in the LAM. The learning performance of the proposed ensemble learning system was evaluated through the software-in-the-loop testing platform (SiL) and hardware-in-the-loop platform (HiL), with the exponential decay method as the baseline for both investigations in SiL and HiL. Four predefined driving cycles were selected to study the learning performance of the proposed method, in which the predefined driving cycle one is used for machine learning. The conclusions drawn from this work are as follows:

(1) In the SiL, under predefined driving cycle one with initial battery SoC 50$\%$, the SBD method performed the best among the three methods with 36.82$\%$ at the end of learning, 1.69$\%$ and 0.72$\%$ higher than the RBD method and the EXD method, respectively. Therefore, the SBD and EXD method is selected as base learners in LAM.

(2) In the SiL, under predefined driving cycle one with initial battery SoC of 50$\%$, the LAM using the weighted-based method outperformed the random-based and maximum-based method with 1.06$\%$ and 5.06$\%$ higher in vehicle energy efficiency. 

(3) In the SiL, under predefined driving cycle one with initial battery SoC of 50$\%$, the investigation of different proportions in the weighted-based method has revealed that the 90$\%$ from the SBD method and 10$\%$ from the EXD method achieved the highest vehicle energy efficiency.

(4) For online real-time control based on the HiL platform under predefined driving cycles 1-4 with different initial battery SoC, the performance of the proposed ensemble learning system with the weighted-based method is robust. It can save more than 1.04$\%$ energy compared to the conventional exponential decay method. 

\section{Acknowledgement}
The work is partially funded by Innovate UK (Grant 102253). It is also sponsored by the State Key Laboratory of Automotive Safety and Energy (Tsinghua University) under Project No. KF2029. The authors gratefully acknowledge the support from Textron Ground Support Equipment UK and Hyper-Drive Innovation.

\bibliographystyle{IEEEtran}
\bibliography{Refs}



\end{document}